\documentclass[11pt,letterpaper]{article}
\usepackage{emnlp2016}
\usepackage{times}
\usepackage{latexsym}
\usepackage{times}
\usepackage{url}
\usepackage{latexsym}
\usepackage{epsfig}
\usepackage{array}
\usepackage{color}
\usepackage{multirow}
\usepackage{url}
\usepackage{subfig}
\usepackage{algorithmic}
\usepackage{algorithm}
\usepackage{multirow}
\usepackage{rotating}
\usepackage{booktabs}
\usepackage{makecell}
\usepackage{comment}
\usepackage{amsmath,amsfonts,amssymb}
\usepackage{tikz}
\usepackage{verbatim}
\usepackage{caption}
\usepackage{booktabs}
\usepackage{ctable}
\usepackage{makecell}
\usepackage{array} % for defining a new column type
\usepackage{varwidth} %for the varwidth minipage environment
\usepackage{stmaryrd}
\usepackage{subfig}
\usepackage{titlesec}

\emnlpfinalcopy

\def\emnlppaperid{***}

\newcommand*{\Scale}[2][4]{\scalebox{#1}{$#2$}}%

\title{Conditional Generation and Snapshot Learning in \\ Neural Dialogue Systems}

\author{Tsung-Hsien Wen, Milica Ga{\v{s}}i\'c, Nikola Mrk{\v{s}}i\'c, Lina M. Rojas-Barahona, \\ {\bf Pei-Hao Su, Stefan Ultes, David Vandyke, Steve Young} \\
  Cambridge University Engineering Department,  \\
  Trumpington Street, 
  Cambridge, CB2 1PZ, UK\\
  {\tt \{thw28,mg436,nm480,lmr46,phs26,su259,djv27,sjy11\}@cam.ac.uk}\
}

\date{}

\begin{document}

\maketitle

\begin{abstract}

Recently a variety of LSTM-based conditional language models (LM) have been applied across a range of language generation tasks.
%However, less effort has been put in studying and interpreting the effectiveness of the various proposed architectures.
In this work we study various model architectures and different ways to represent and aggregate the source information in an end-to-end neural dialogue system framework.
A method called snapshot learning is also proposed to facilitate learning from supervised sequential signals by applying a {\it companion} cross-entropy objective function to the conditioning vector.
The experimental and analytical results demonstrate firstly that competition occurs between the conditioning vector and the LM, and the differing architectures provide different trade-offs between the two.
Secondly, the discriminative power and transparency of the conditioning vector is key to providing both model interpretability and better performance.  Thirdly, snapshot learning leads to consistent performance improvements independent of which architecture is used.

\end{abstract}

\section{Introduction}\label{sec:intro}

Recurrent Neural Network (RNN)-based conditional language models (LM) have been shown to be very effective in solving a number of real world problems, such as machine translation (MT)~\cite{ChoMGBSB14} and image caption generation~\cite{KarpathyF14}.
Recently, RNNs were applied to task of generating sentences from an explicit semantic representation~\cite{wenrgm15}.
Attention-based methods~\cite{ENCDEC_GEN15} and Long Short-term Memory (LSTM)-like~\cite{Hochreiter1997} gating mechanisms~\cite{wensclstm15} have both been studied to improve generation quality.
Although it is now clear that LSTM-based conditional LMs can generate plausible natural language, less effort has been put in comparing the different  model architectures.
Furthermore, conditional generation models are typically tested on relatively straightforward tasks conditioned on a single source (e.g. a sentence or an image) and where the goal is to optimise a single metric (e.g. BLEU).
In this work, we study the use of conditional LSTMs in the generation component of neural network (NN)-based dialogue systems which depend on multiple conditioning sources and optimising multiple metrics.

Neural conversational agents~\cite{VinyalsL15,ShangLL15} are direct extensions of the sequence-to-sequence model~\cite{SutskeverVL14} in which a conversation is cast as a source to target transduction problem.
However, these models are still far from real world applications because they lack any capability for supporting domain specific tasks, for example, being able to interact with databases~\cite{Sukhbaatar2015EndToEndMN,YinLLK15} and aggregate useful information into their responses.
Recent work by~\newcite{wenn2ndialog16}, however, proposed an end-to-end trainable neural dialogue system that can assist users to complete specific tasks.
Their system used both distributed and symbolic representations to capture user intents, and these collectively condition a NN language generator to generate system responses.
Due to the diversity of the conditioning information sources, the best way to represent and combine them is non-trivial.

In~\newcite{wenn2ndialog16}, the objective function for learning the dialogue policy and language generator depends solely on the likelihood of the output sentences.
However, this sequential supervision signal may not be informative enough to learn a good conditioning vector representation resulting in a generation process which is dominated by the LM.  This can often lead to inappropriate system outputs.

In this paper, we therefore also investigate the use of snapshot learning which attempts to mitigate this problem by heuristically applying {\it companion} supervision signals to a subset of the conditioning vector.
This idea is similar to deeply supervised nets~\cite{deepsuplee15} in which the final cost from the output layer is optimised together with the companion signals generated from each intermediary layer.
We have found that snapshot learning offers several benefits: (1) it consistently improves performance; (2) it learns discriminative and robust feature representations and alleviates the vanishing gradient problem; (3) it appears to learn transparent and interpretable subspaces of the conditioning vector.

\section{Related Work}\label{sec:related}

Machine learning approaches to task-oriented dialogue system design have cast the problem as a partially observable Markov Decision Process (POMDP)~\cite{6407655} with the aim of using reinforcement learning (RL) to train dialogue policies online through interactions with real users~\cite{6639297}.
In order to make RL tractable, the state and action space must be carefully designed~\cite{Young2010HIS} and the understanding~\cite{henderson14,Mrksic15} and generation~\cite{wensclstm15,wenmultinlg16} modules were assumed available or trained standalone on supervised corpora.
Due to the underlying hand-coded semantic representation~\cite{Traum1999}, the conversation is far from natural and the comprehension capability is limited.
This motivates the use of neural networks to model dialogues from end to end as a conditional generation problem.

Interest in generating natural language using NNs can be attributed to the success of RNN LMs for large vocabulary speech recognition~\cite{39298195,5947611}.
~\newcite{ICML2011Sutskever_524} showed that plausible sentences can be obtained by sampling characters one by one from the output layer of an RNN.
By conditioning an LSTM on a sequence of characters,~\newcite{Graves13} showed that machines can synthesise handwriting indistinguishable from that of a human.
Later on, this conditional generation idea has been tried in several research fields, for example, generating image captions by conditioning an RNN on a convolutional neural network (CNN) output~\cite{KarpathyF14,XuBKCCSZB15}; 
translating a source to a target language by conditioning a decoder LSTM on top of an encoder LSTM~\cite{ChoMGBSB14,BahdanauCB14};
or generating natural language by conditioning on a symbolic semantic representation~\cite{wensclstm15,ENCDEC_GEN15}.
Among all these methods, attention-based mechanisms~\cite{BahdanauCB14,HermannKGEKSB15,wangling16lpn} have been shown to be very effective improving performance using a dynamic source aggregation strategy.

To model dialogue as conditional generation, a sequence-to-sequence learning~\cite{SutskeverVL14} framework has been adopted.
~\newcite{VinyalsL15} trained the same model on several conversation datasets and showed that the model can generate plausible conversations.
However,~\newcite{SerbanSBCP15} discovered that the majority of the generated responses are generic due to the maximum likelihood criterion, which was latter addressed by~\newcite{LiGBGD15} using a maximum mutual information decoding strategy. 
Furthermore, the lack of a consistent system persona was also studied in~\newcite{Lipersona16}.
Despite its demonstrated potential, a major barrier for this line of research is data collection.
Many works~\cite{ryanubuntu15,SerbanLCP15,DodgeGZBCMSW15} have investigated conversation datasets for developing chat bot or QA-like general purpose conversation agents.
However, collecting data to develop goal oriented dialogue systems that can help users to complete a task in a specific domain remains difficult.
In a recent work by~\newcite{wenn2ndialog16}, this problem was addressed by designing an online, parallel version of Wizard-of-Oz data collection~\cite{Kelley84} which allows large scale and cheap in-domain conversation data to be collected using Amazon Mechanical Turk.
An NN-based dialogue model was also proposed to learn from the collected dataset and was shown to be able to assist human subjects to complete specific tasks.

Snapshot learning can be viewed as a special form of weak supervision (also known as distant- or self supervision)~\cite{Craven1999,ilprints665}, in which supervision signals are heuristically labelled by matching unlabelled corpora with entities or attributes in a structured database.
It has been widely applied to relation extraction~\cite{Mintz2009} and information extraction~\cite{Hoffmann2011} in which facts from a knowledge base (e.g. Freebase) were used as objectives to train classifiers. 
Recently, self supervision was also used in memory networks~\cite{hill2016} to improve the discriminative power of memory attention.
Conceptually, snapshot learning is related to curriculum learning~\cite{Bengio2009}.
Instead of learning easier examples before difficult ones, snapshot learning creates an easier target for each example.
In practice, snapshot learning is similar to deeply supervised nets~\cite{deepsuplee15} in which {\it companion} objectives are generated from intermediary layers and optimised altogether with the output objective.

\section{Neural Dialogue System}\label{sec:ndial}

The testbed for this work is a neural network-based task-oriented dialogue system proposed by~\newcite{wenn2ndialog16}. %, as shown in Figure~\ref{fig:n2n}.
The model casts dialogue as a source to target sequence transduction problem (modelled by a sequence-to-sequence architecture~\cite{SutskeverVL14}) augmented with the dialogue history (modelled by a belief tracker~\cite{henderson14}) and the current database search outcome (modelled by a database operator).
The model consists of both encoder and decoder modules. 
The details of each module are given below.

%\begin{figure*}[t]
%\centerline{\includegraphics[width=110mm]{n2n}}
%\caption{The neural dialogue system framework}
%\label{fig:n2n}
%\end{figure*}

\subsection{Encoder Module}\label{ssec:enc}

At each turn $t$, the goal of the encoder is to produce a distributed representation of the system action $\mathrm{\mathbf{m}}_{t}$, which is then used to condition a decoder to generate the next system response in skeletal form\footnote{\label{fn:delex}Delexicalisation: slots and values are replaced by generic tokens (e.g. keywords like {\it Chinese food} are replaced by {\it [v.food] [s.food]} to allow weight sharing.}.
It consists of four submodules: intent network, belief tracker, database operator, and policy network.

\noindent{\bf Intent Network} \hspace{2mm} The intent network takes a sequence of tokens\textsuperscript{\ref{fn:delex}} and converts it into a sentence embedding representing the user intent using an LSTM network. 
The hidden layer of the LSTM at the last encoding step $\mathrm{\mathbf{z}}_t$ is taken as the representation.
As mentioned in ~\newcite{wenn2ndialog16}, this representation can be viewed as a distributed version of the speech act~\cite{Traum1999} used in traditional systems.

\noindent{\bf Belief Trackers} \hspace{2mm} In addition to the intent network, the neural dialogue system uses a set of slot-based belief trackers~\cite{henderson14,Mrksic15} to track user requests. 
By taking each user input as new evidence, the task of a belief tracker is to maintain a multinomial distribution $p$ over values $v \in V_{s}$ for each informable slot $s$\footnote{\label{fn:slots}Informable slots are slots that users can use to constrain the search, such as food type or price range; Requestable slots are slots that users can ask a value for, such as phone number. This information is specified in the domain ontology.}, and a binary distribution for each requestable slot\textsuperscript{\ref{fn:slots}}.
These probability distributions $\mathrm{\mathbf{p}}^s_{t}$ are called belief states of the system.
The belief states $\mathrm{\mathbf{p}}_t^{s}$, together with the intent vector $\mathrm{\mathbf{z}}_t$, can be viewed as the system's comprehension of the user requests up to turn $t$.

\begin{figure*}[t]
\vspace{-3mm}
\subfloat[Language model type LSTM]{\includegraphics[width=4.5cm]{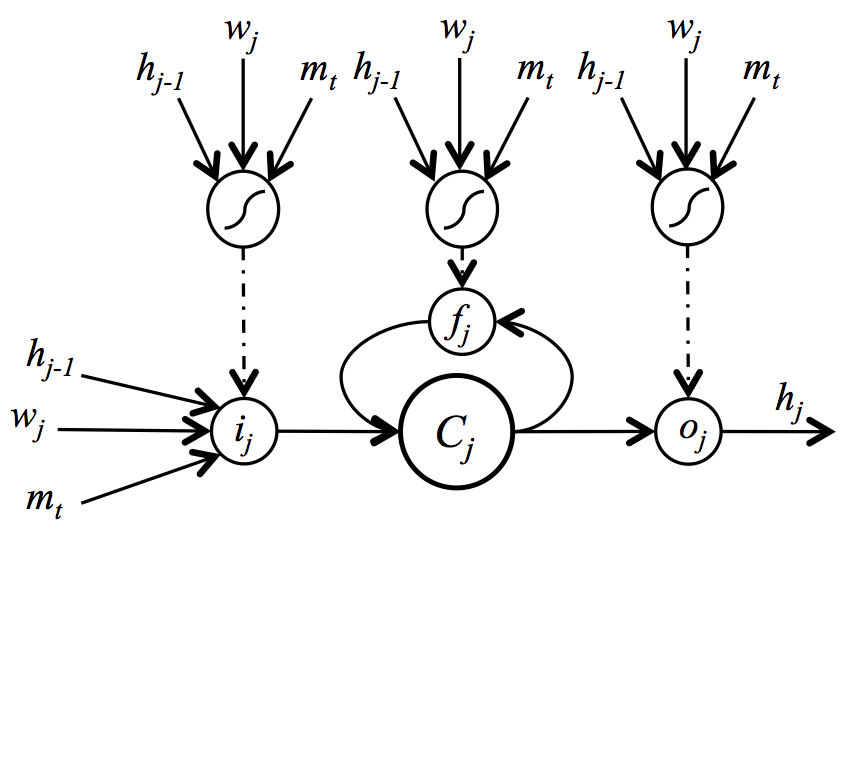}\label{sfig:lm}}
\hfill
\subfloat[Memory type LSTM]{\includegraphics[width=4.5cm]{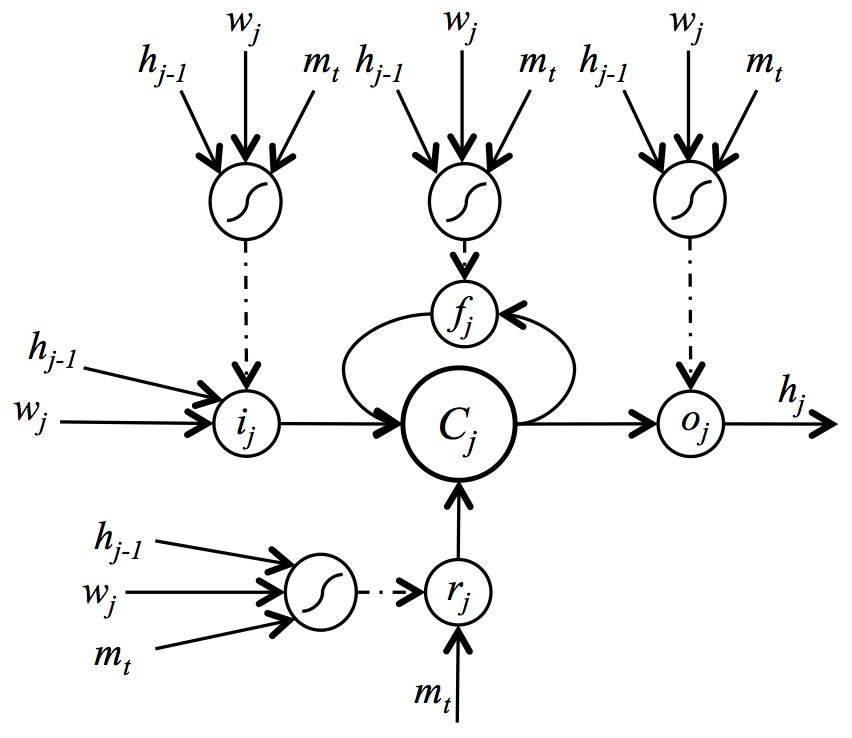}\label{sfig:mem}}
\hfill
\subfloat[Hybrid type LSTM]{\includegraphics[width=4.8cm]{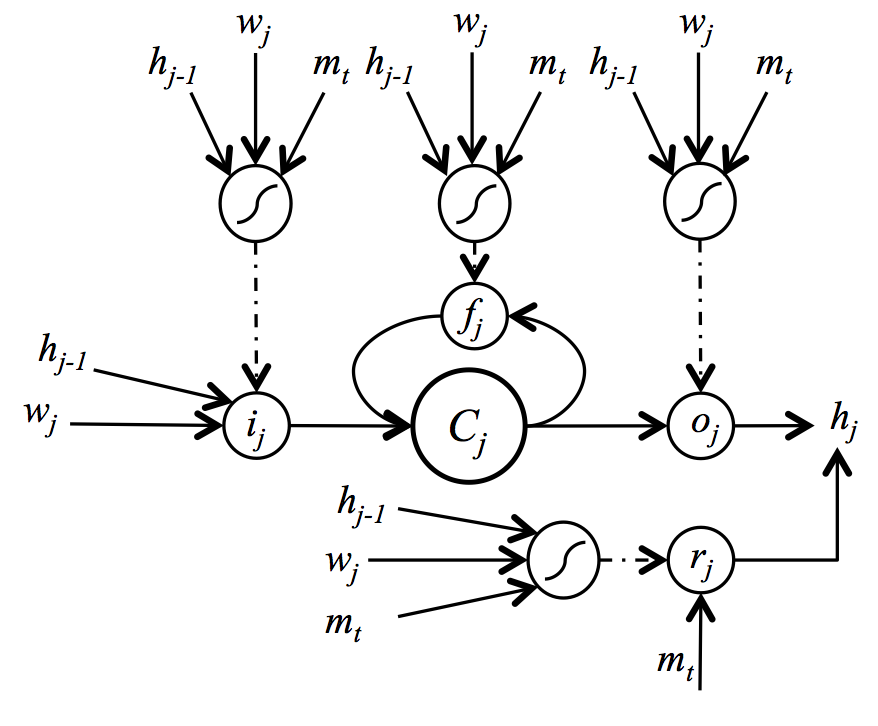}\label{sfig:hybrid}}
\caption{Three different conditional generation architectures.}
\vspace{-3mm}
\end{figure*}

\noindent{\bf Database Operator} \hspace{2mm} Based on the belief states $\mathrm{\mathbf{p}}_t^{s}$, a DB query is formed by taking the union of the maximum values of each informable slot.
A vector $\mathrm{\mathbf{x}}_t$ representing different degrees of matching in the DB (no match, 1 match, ... or  more than 5 matches) is produced by counting the number of matched entities and expressing it as a 6-bin 1-hot encoding.
If $\mathrm{\mathbf{x}}_t$ is not zero, an associated entity pointer is maintained which identifies one of the matching DB entities selected at random. The entity pointer is updated if the current entity no longer matches the search criteria; otherwise it stays the same.

\noindent{\bf Policy Network} \hspace{2mm} Based on the vectors $\mathrm{\mathbf{z}}_t$, $\mathrm{\mathbf{p}}_t^s$, and $\mathrm{\mathbf{x}}_t$ from the above three modules, the policy network combines them into a single action vector $\mathrm{\mathbf{m}}_t$ by a three-way matrix transformation, 
\begin{equation}\label{eq:mt}
\Scale[0.85]{\mathrm{\mathbf{m}}_t = \tanh(\mathrm{\mathbf{W}}_{zm}\mathrm{\mathbf{z}}_{t} + \mathrm{\mathbf{W}}_{xm}\mathrm{\mathbf{x}}_{t} + \sum_{s\in \mathbb{G}}\mathrm{\mathbf{W}}^{s}_{pm}\mathrm{\mathbf{p}}_{t}^s)}
\end{equation}
where matrices $\mathrm{\mathbf{W}}_{zm}$, $\mathrm{\mathbf{W}}_{pm}^s$, and $\mathrm{\mathbf{W}}_{xm}$ are parameters and $\mathbb{G}$ is the domain ontology.

\subsection{Decoder Module}\label{ssec:dec}

Conditioned on the system action vector $\mathrm{\mathbf{m}}_t$ provided by the encoder module, the decoder module uses a conditional LSTM LM to generate the required system output token by token in skeletal form\textsuperscript{\ref{fn:delex}}. 
The final system response can then be formed by substituting the actual values of the database entries into the skeletal sentence structure.

\subsubsection{Conditional Generation Network}\label{sssec:condgen}

\noindent In this paper we study and analyse three different variants of LSTM-based conditional generation architectures:

\noindent {\bf Language Model Type} \hspace{2mm}
The most straightforward way to condition the LSTM network on additional source information is to concatenate the conditioning vector $\mathrm{\mathbf{m}}_t$ together with the input word embedding $\mathrm{\mathbf{w}}_j$ and previous hidden layer $\mathrm{\mathbf{h}}_{j-1}$,
\[
 	\Scale[0.8]{\left(\begin{array}{ll}
		\mathrm{\mathbf{i}}_j\\
		\mathrm{\mathbf{f}}_j\\
		\mathrm{\mathbf{o}}_j\\
		\hat{\mathrm{\mathbf{c}}}_j
	\end{array}\right)
	=
	\left(\begin{array}{ll}
		\text{sigmoid}\\
		\text{sigmoid}\\
		\text{sigmoid}\\
		\tanh
	\end{array}\right)
	\mathrm{\mathbf{W}}_{4n,3n}
	\left(\begin{array}{ll}
		\mathrm{\mathbf{m}}_{t}\\
		\mathrm{\mathbf{w}}_j\\
		\mathrm{\mathbf{h}}_{j-1}
	\end{array}\right)}
\]
\[
	\Scale[0.8]{\mathrm{\mathbf{c}}_j = \mathrm{\mathbf{f}}_{j} \odot \mathrm{\mathbf{c}}_{j-1} + \mathrm{\mathbf{i}}_{j} \odot \hat{\mathrm{\mathbf{c}}}_j}
\]
\[
	\Scale[0.8]{\mathrm{\mathbf{h}}_j =  \mathrm{\mathbf{o}}_{j} \odot \tanh(\mathrm{\mathbf{c}}_j)}
\]
where index $j$ is the generation step, $n$ is the hidden layer size, $\mathrm{\mathbf{i}}_j,\mathrm{\mathbf{f}}_j,\mathrm{\mathbf{o}}_j \in [0,1]^n$ are input, forget, and output gates respectively, $\hat{\mathrm{\mathbf{c}}}_j$ and $\mathrm{\mathbf{c}}_j$ are proposed cell value and true cell value at step $j$,
and $\mathrm{\mathbf{W}}_{4n,3n}$ are the model parameters.
The model is shown in Figure~\ref{sfig:lm}.
Since it does not differ significantly from the original LSTM, we call it the {\it language model type} (lm) conditional generation network.

\noindent{\bf Memory Type} \hspace{2mm} The {\it memory type} (mem) conditional generation network was introduced by~\newcite{wensclstm15}, shown in Figure~\ref{sfig:mem}, in which the conditioning vector $\mathrm{\mathbf{m}}_{t}$ is governed by a standalone reading gate $\mathrm{\mathbf{r}}_{j}$. 
This reading gate decides how much information should be read from the conditioning vector and directly writes it into the memory cell $\mathrm{\mathbf{c}}_{j}$,
\[
 	\Scale[0.8]{\left(\begin{array}{ll}
		\mathrm{\mathbf{i}}_j\\
		\mathrm{\mathbf{f}}_j\\
		\mathrm{\mathbf{o}}_j\\
		\mathrm{\mathbf{r}}_j
	\end{array}\right)
	=
	\left(\begin{array}{ll}
		\text{sigmoid}\\
		\text{sigmoid}\\
		\text{sigmoid}\\
		\text{sigmoid}\\
	\end{array}\right)
	\mathrm{\mathbf{W}}_{4n,3n}
	\left(\begin{array}{ll}
		\mathrm{\mathbf{m}}_{t}\\
		\mathrm{\mathbf{w}}_j\\
		\mathrm{\mathbf{h}}_{j-1}
	\end{array}\right)}
\]
\[
	\Scale[0.8]{\hat{\mathrm{\mathbf{c}}}_j = \tanh\big(\mathrm{\mathbf{W}}_c(\mathrm{\mathbf{w}}_j\oplus\mathrm{\mathbf{h}}_{j-1})\big)}
\]
\[
	\Scale[0.8]{\mathrm{\mathbf{c}}_j = \mathrm{\mathbf{f}}_{j} \odot \mathrm{\mathbf{c}}_{j-1} + \mathrm{\mathbf{i}}_{j} \odot \hat{\mathrm{\mathbf{c}}}_j + \mathrm{\mathbf{r}}_{j} \odot \mathrm{\mathbf{m}}_t}
\]
\[
	\Scale[0.8]{\mathrm{\mathbf{h}}_j =  \mathrm{\mathbf{o}}_{j} \odot \tanh(\mathrm{\mathbf{c}}_j)}
\]
where $\mathrm{\mathbf{W}}_{c}$ is another weight matrix to learn. 
The idea behind this is that the model isolates the conditioning vector from the LM so that the model has more flexibility to learn to trade off between the two.

\noindent{\bf Hybrid Type} \hspace{2mm} Continuing with the same idea as the {\it memory type} network, a complete separation of conditioning vector and LM (except for the gate controlling the signals) is provided by the {\it hybrid type} network shown in Figure~\ref{sfig:hybrid},
\[
 	\Scale[0.8]{\left(\begin{array}{ll}
		\mathrm{\mathbf{i}}_j\\
		\mathrm{\mathbf{f}}_j\\
		\mathrm{\mathbf{o}}_j\\
		\mathrm{\mathbf{r}}_j
	\end{array}\right)
	=
	\left(\begin{array}{ll}
		\text{sigmoid}\\
		\text{sigmoid}\\
		\text{sigmoid}\\
		\text{sigmoid}
	\end{array}\right)
	\mathrm{\mathbf{W}}_{4n,3n}
	\left(\begin{array}{ll}
		\mathrm{\mathbf{m}}_{t}\\
		\mathrm{\mathbf{w}}_j\\
		\mathrm{\mathbf{h}}_{j-1}
	\end{array}\right)}
\]
\[
	\Scale[0.8]{\hat{\mathrm{\mathbf{c}}}_j = \tanh\big(\mathrm{\mathbf{W}}_c(\mathrm{\mathbf{w}}_j\oplus\mathrm{\mathbf{h}}_{j-1})\big)}
\]
\[
	\Scale[0.8]{\mathrm{\mathbf{c}}_j = \mathrm{\mathbf{f}}_{j} \odot \mathrm{\mathbf{c}}_{j-1} + \mathrm{\mathbf{i}}_{j} \odot \hat{\mathrm{\mathbf{c}}}_j}
\]
\[
	\Scale[0.8]{\mathrm{\mathbf{h}}_j =  \mathrm{\mathbf{o}}_{j} \odot \tanh(\mathrm{\mathbf{c}}_j)+ \mathrm{\mathbf{r}}_{j} \odot \mathrm{\mathbf{m}}_t}
\]
This model was motivated by the fact that long-term dependency is not needed for the conditioning vector because we apply this information at every step $j$ anyway.
The decoupling of the conditioning vector and the LM is attractive because it leads to better interpretability of the results and provides the potential to learn a better conditioning vector and LM.

\subsubsection{Attention and Belief Representation}\label{sssec:att+bef}
\noindent {\bf Attention} \hspace{1.5mm} An attention-based mechanism provides an effective approach for aggregating multiple information sources for prediction tasks.
Like~\newcite{wenn2ndialog16}, we explore the use of an attention mechanism to combine the tracker belief states in which the policy network in Equation~\ref{eq:mt} is modified as 
\begin{equation*}
\Scale[0.85]{\mathrm{\mathbf{m}}_t^{j} = \tanh(\mathrm{\mathbf{W}}_{zm}\mathrm{\mathbf{z}}_{t} +\mathrm{\mathbf{W}}_{xm}\mathrm{\mathbf{x}}_{t}+ \sum_{s\in \mathbb{G}}\alpha_{s}^{j}\mathrm{\mathbf{W}}^{s}_{pm}\mathrm{\mathbf{p}}_{t}^s)}
\end{equation*}
where the attention weights $\alpha_{s}^{j}$ are calculated by, 
\begin{equation*}
\Scale[0.85]{\alpha_{s}^{j} = \text{softmax}\big( \mathrm{\mathbf{r}}^\intercal \tanh\big( \mathrm{\mathbf{W}}_{r} \cdot  
(   \mathrm{\mathbf{v}}_{t}\oplus \mathrm{\mathbf{p}}_{t}^{s} \oplus \mathrm{\mathbf{w}}_{j}^{t} \oplus \mathrm{\mathbf{h}}_{j-1}^{t} ) \big) \big)}
\end{equation*}
where $\mathrm{\mathbf{v}}_{t}=\mathrm{\mathbf{z}}_{t}+\mathrm{\mathbf{x}}_{t}$ and matrix $\mathrm{\mathbf{W}}_{r}$ and vector $\mathrm{\mathbf{r}}$ are parameters to learn.

\noindent {\bf Belief Representation} \hspace{2mm} The effect of different belief state representations on the end performance are also studied.
For user informable slots, the {\it full} belief state $\mathrm{\mathbf{p}}_{t}^{s}$ is the original state containing all categorical values;
the {\it summary} belief state contains only three components: the summed value of all categorical probabilities, the probability that the user said they ``don't care" about this slot and the probability that the slot has not been mentioned.
For user requestable slots, on the other hand, the full belief state is the same as the summary belief state because the slot values are binary rather than categorical.

\subsection{Snapshot Learning}\label{ssec:snapshot}

Learning conditional generation models from sequential supervision signals can be difficult, because it requires the model to learn both long-term word dependencies and potentially distant source encoding functions.
To mitigate this difficulty, we introduce a novel method called {\it snapshot learning} to create a vector of binary labels $\mathrm{\mathbf{\Upsilon}}^{j}_{t} \in [0,1]^{d}$, $d<dim(\mathrm{\mathbf{m}}_{t}^{j})$ as the {\it snapshot} of the remaining part of the output sentence $T_{t,j:|T_t|}$ from generation step $j$.
Each element of the snapshot vector is an indicator function of a certain event that will happen in the future, which can be obtained either from the system response or dialogue context at training time.
A {\it companion} cross entropy error is then computed to force a subset of the conditioning vector $\mathrm{\mathbf{\hat{m}}}_{t}^{j} \subset \mathrm{\mathbf{m}}_{t}^{j}$ to be close to the snapshot vector,
\begin{equation}\label{eq:snapshot}
\Scale[0.85]{L_{ss}(\cdot) = -\sum_{t}\sum_{j} \mathop{\mathbb{E}}[H(\Upsilon_{t}^{j}, \hat{m}_{t}^{j})]}
\end{equation}
where $H(\cdot)$ is the cross entropy function, $\Upsilon_{t}^{j}$ and $\hat{m}_{t}^{j}$ are elements of vectors $\mathrm{\mathbf{\Upsilon}}^{j}_{t}$ and $\mathrm{\mathbf{\hat{m}}}_{t}^{j}$, respectively. 
In order to make the $\tanh$ activations of $\mathrm{\mathbf{\hat{m}}}_{t}^{j}$ compatible with the 0-1 snapshot labels, we squeeze each value of $\mathrm{\mathbf{\hat{m}}}_{t}^{j}$ by adding 1 and dividing by 2 before computing the cost.

\begin{figure}[t]
\centerline{\includegraphics[width=60mm]{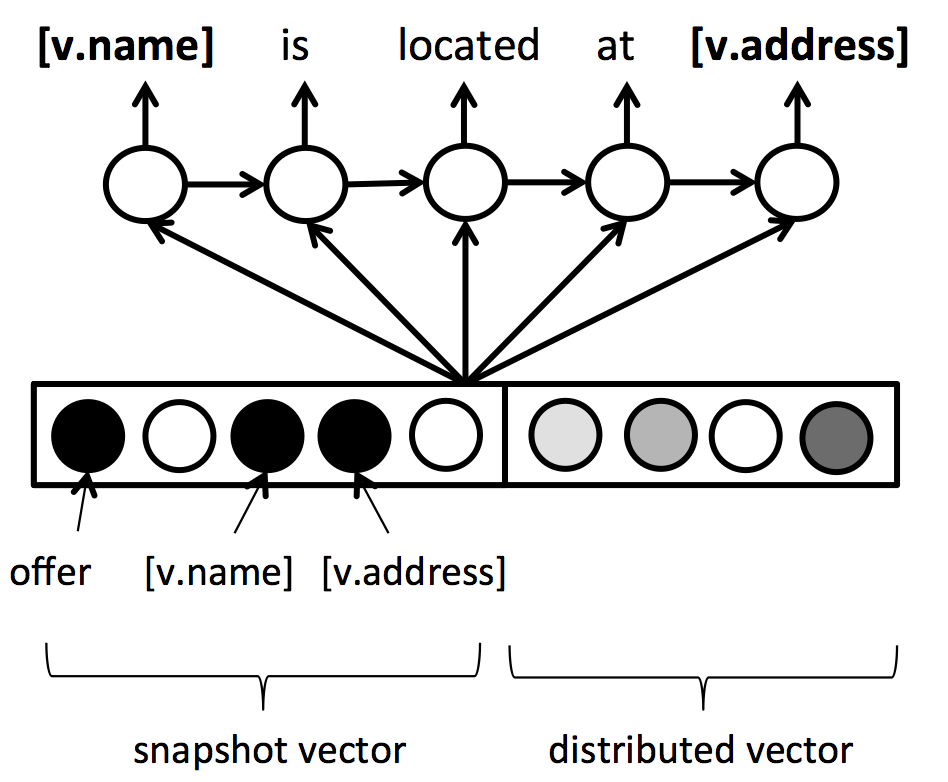}}
\caption{The idea of snapshot learning. The snapshot vector was trained with additional supervisions on a set of indicator functions heuristically labelled using the system response.}
\label{fig:snapshot}
\vspace{-4mm}
\end{figure}

\begin{table*}[t]
  \centering
  \Scale[0.95]{\begin{tabular}{llllcccc}
    \toprule
    &		Architecture 	&	Belief	&	Success(\%)	&	SlotMatch(\%)	&	T5-BLEU	&	T1-BLEU\\
    \midrule
    \multicolumn{7}{l}{\bf Belief state representation}\\
    &		lm		&	full	 	&  	72.6 / 74.5	&	52.1 / 60.3   	& 	0.207 / 0.229   	&	0.216 / 0.238 \\
    &		lm		&	summary  & 	74.5	/ 76.5	&	57.4 / 61.2	& 	0.221 / 0.231    	&	0.227 / 0.240 \\
% &		lm		&	simplified  & 	76.5	/ 76.3	&	61.3 / 61.1	& 	0.230 / 0.235   	&	0.240 / 0.245\\
    \midrule
    \multicolumn{7}{l}{\bf Conditional architecture}\\
    &		lm		&	summary  & 	74.5	/ 76.5	&	57.4 / 61.2	& 	0.221 / 0.231   	&	0.227 / 0.240\\
    &   	mem		&	summary	&	75.5 / 77.5	&	59.2 / {\bf 61.3}	&	0.222 / {\bf0.232} 	&	0.231 / {\bf0.243} \\
    &   	hybrid	&	summary	&	76.1 / 79.2	&	52.4 / 60.6	&	0.202 / 0.228 	&	0.212 / 0.237 \\
    \midrule
    %\multicolumn{7}{l}{\bf Attention w/ simplified belief}\\
    %&		lm		&	simplified  & 	78.5	/ 76.8	&	59.9 / 61.6	& 	0.229 / 0.236   	&	0.239 / 0.244\\
    %&   	cond.	&	simplified	&	77.9 / 78.8	&	59.5 / 61.4	&	0.226 / 0.232 	&	0.239 / 0.244 \\
    %&   	mix		&	simplified	&	78.3 / 79.8	&	59.4 / 60.7	&	0.224 / 0.229 	&	0.236 / 0.240 \\
    %\midrule
    \multicolumn{7}{l}{\bf Attention-based model}\\
    &		lm		&	summary  & 	79.4	/ 78.2	&	60.6 / 60.2	& 	0.228 / 0.231   	&	0.239 / 0.241\\
    &   	mem		&	summary	&	76.5 / 80.2	&	57.4 / 61.0	&	0.220 / 0.229 	&	0.228 / 0.239 \\
    &   	hybrid	&	summary	&	79.0 / {\bf81.8}	&	56.2 / 60.5	&	0.214 / 0.227 	&	0.224 / 0.240 \\
    \bottomrule
  \end{tabular}}
  \caption{Performance comparison of different model architectures, belief state representations, and snapshot learning. The numbers to the left and right of the {\bf /} sign are learning without and with snapshot, respectively. The model with the best performance on a particular metric (column) is shown in bold face.}
  \label{lab:exp}
  \vspace{-3mm}
\end{table*}

The indicator functions we use in this work have two forms: (1) whether a particular slot value (e.g., {\it[v.food]}\textsuperscript{\ref{fn:delex}}) is going to occur, and (2) whether the system has offered a venue\footnote{Details of the specific application used in this study
are given in Section~\ref{sec:exp} below.}, as shown in Figure~\ref{fig:snapshot}.
The {\it offer} label in the snapshot is produced by checking the delexicalised name token ({\it[v.name]}) in the entire dialogue. 
If it has occurred, every label in subsequent turns is labelled with 1. Otherwise it is labelled with 0.
To create snapshot targets for a particular slot value, the output sentence is matched with the corresponding delexicalised token turn by turn, per generation step.
At each generation step, the target is labelled with 0 if that delexicalised token has been generated; otherwise it is set to 1.
However, for the models without attention, the targets per turn are set to the same because the condition vector will not be able to learn the dynamically changing behaviour without attention.

\section{Experiments}\label{sec:exp}

{\bf Dataset} \hspace{2mm} The dataset used in this work was collected in the Wizard-of-Oz online data collection described by~\newcite{wenn2ndialog16}, in which the task of the system is to assist users to find a restaurant in Cambridge, UK area.
There are three informable slots ({\it food}, {\it pricerange}, {\it area}) that users can use to constrain the search and six requestable slots ({\it address}, {\it phone}, {\it postcode} plus the three informable slots) that the user can ask a value for once a restaurant has been offered.
There are 676 dialogues in the dataset (including both finished and unfinished dialogues) and approximately 2750 turns in total.
The database contains 99 unique restaurants.

\begin{figure*}
\vspace{-4mm}
\centering
\hfill
\subfloat[{\it Hybrid} LSTM w/o snapshot learning]{\includegraphics[width=5.5cm]{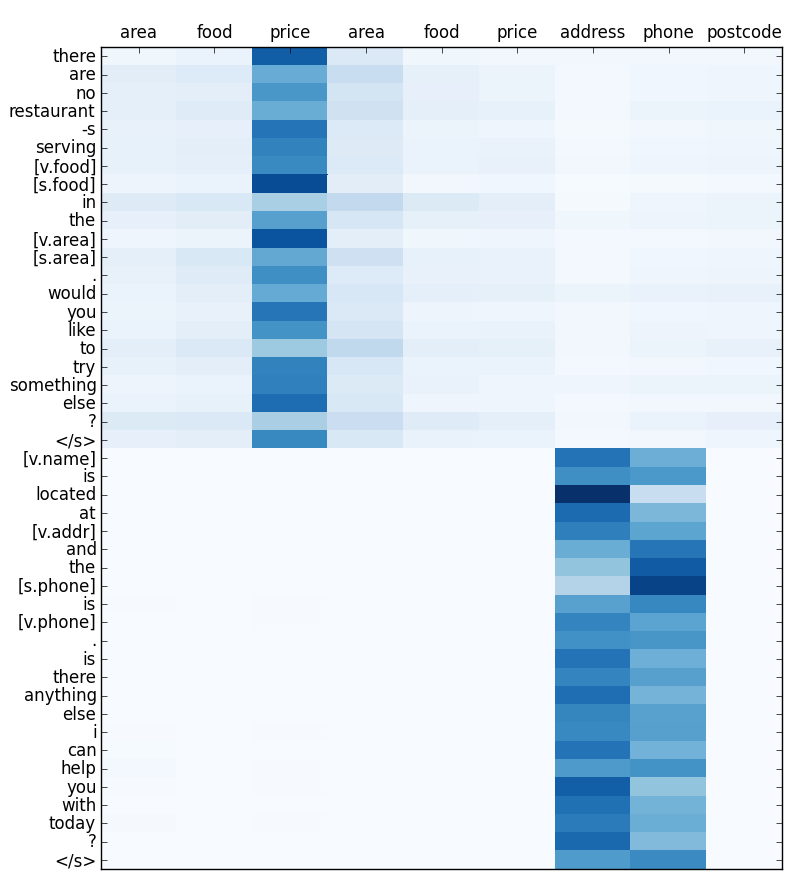}\label{fig:att1}}
\hfill
\subfloat[{\it Hybrid} LSTM w/ snapshot learning]{\includegraphics[width=5.5cm]{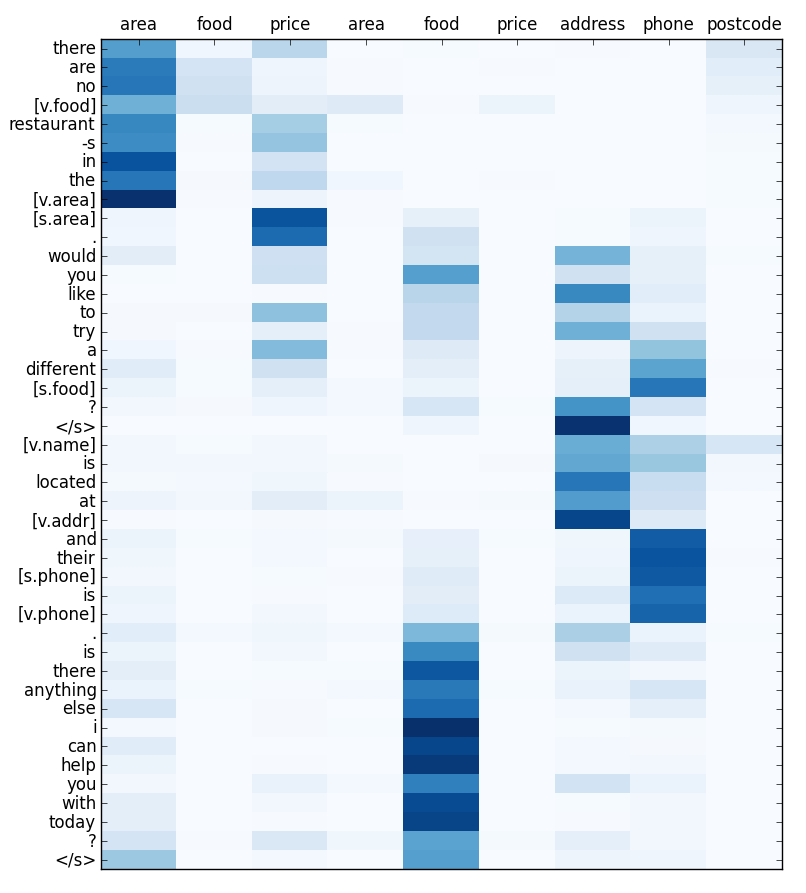}\label{fig:att2}}
\hfill\hfill
\caption{Learned attention heat maps over trackers. The first three columns in each figure are informable slot trackers and the rest are requestable slot trackers. The generation model is the {\it hybrid} type LSTM.}
\label{fig:att}
\vspace{-3mm}
\end{figure*}

\noindent {\bf Training} \hspace{2mm} The training procedure was divided into two stages. 
Firstly, the belief tracker parameters $\theta_{b}$ were pre-trained using cross entropy errors between tracker labels and predictions.
Having fixed the tracker parameters, the remaining parts of the model $\theta_{\backslash b}$ are trained using the cross entropy errors from the generation network LM,
\begin{equation}
\Scale[0.85]{L(\theta_{\backslash b}) = -\sum_{t}\sum_{j} H(\mathrm{\mathbf{y}}_{j}^{t},\mathrm{\mathbf{p}}_{j}^{t}) + \lambda L_{ss}(\cdot)}
\end{equation}
where $\mathrm{\mathbf{y}}_{j}^{t}$ and $\mathrm{\mathbf{p}}_{j}^{t}$ are output token targets and predictions respectively, at turn $t$ of output step $j$,
 $L_{ss}(\cdot)$ is the snapshot cost from Equation~\ref{eq:snapshot}, and $\lambda$ is the tradeoff parameter in which we set to 1 for all models trained with snapshot learning.
We treated each dialogue as a batch and used stochastic gradient descent with a small $l2$ regularisation term to train the model.
The collected corpus was partitioned into a training, validation, and testing sets in the ratio 3:1:1. 
Early stopping was implemented based on the validation set considering only LM log-likelihoods.
Gradient clipping was set to 1.
The hidden layer sizes were set to 50, and the weights were randomly initialised between -0.3 and 0.3 including word embeddings.
The vocabulary size is around 500 for both input and output, in which rare words and words that can be delexicalised have been removed.

\noindent {\bf Decoding} \hspace{2mm} In order to compare models trained with different recipes rather than decoding strategies, we decode all the trained models with the average log probability of tokens in the sentence.
We applied beam search with a beamwidth equal to 10, the search stops when an end-of-sentence token is generated.
In order to consider language variability, we ran decoding until 5 candidates were obtained and performed evaluation on them.

\noindent {\bf Metrics} \hspace{2mm} We compared models trained with different recipes by performing a corpus-based evaluation in which the model is used to predict each system response in the held-out test set.
Three evaluation metrics were used: BLEU score (on top-1 and top-5 candidates)~\cite{papineni2002bleu}, slot matching rate and objective task success rate~\cite{SuVGKMWY15}.
The dialogue is marked as successful if both: (1) the offered entity matches the task that was specified to the user, and (2) the system answered all the associated information requests (e.g. {\it what is the address?}) from the user.
The slot matching rate is the percentage of delexicalised tokens (e.g. {\it[s.food]} and {\it[v.area]}\textsuperscript{\ref{fn:delex}}) appear in the candidate also appear in the reference.
We computed the BLEU scores on the skeletal sentence forms before substituting with the actual entity values. 
All the results were averaged over 10 random initialised networks. 

\captionsetup[subfigure]{position=top, labelfont=bf,textfont=normalfont,singlelinecheck=off,justification=raggedright}
\begin{figure*}[t]
\vspace{-1mm}
\centering
\raisebox{\dimexpr-.5\height+1em}{\includegraphics[width=120mm]{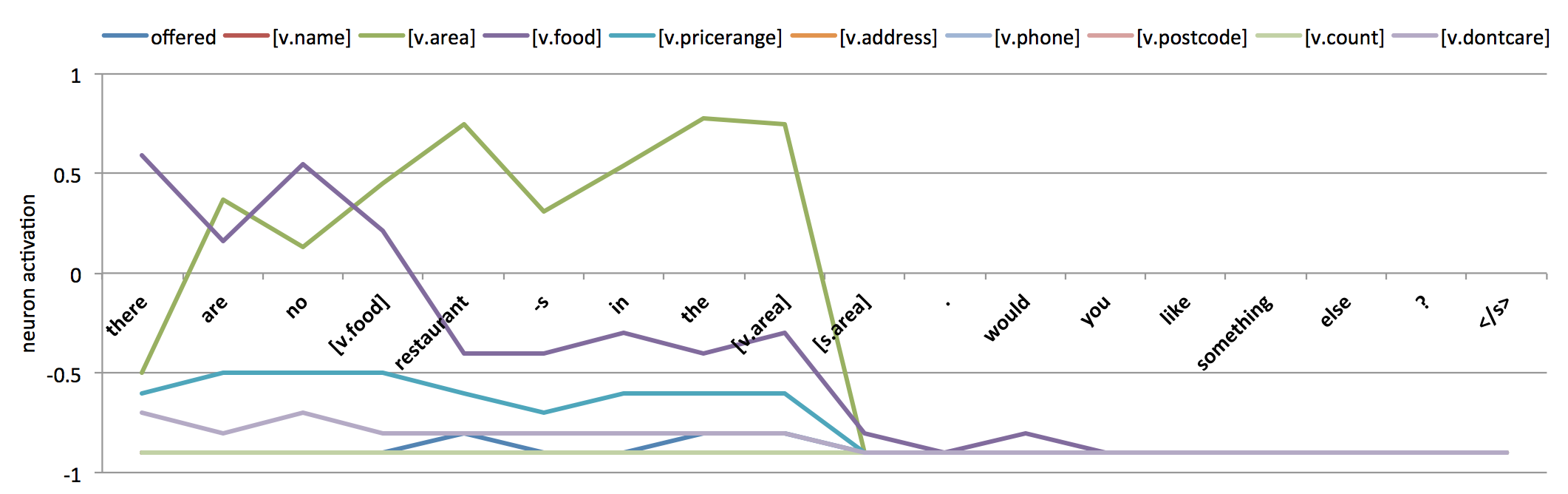}}\ \subfloat[\label{fig:snap1}]{} \\[\topskip]
\raisebox{\dimexpr-.5\height+1em}{\includegraphics[width=120mm]{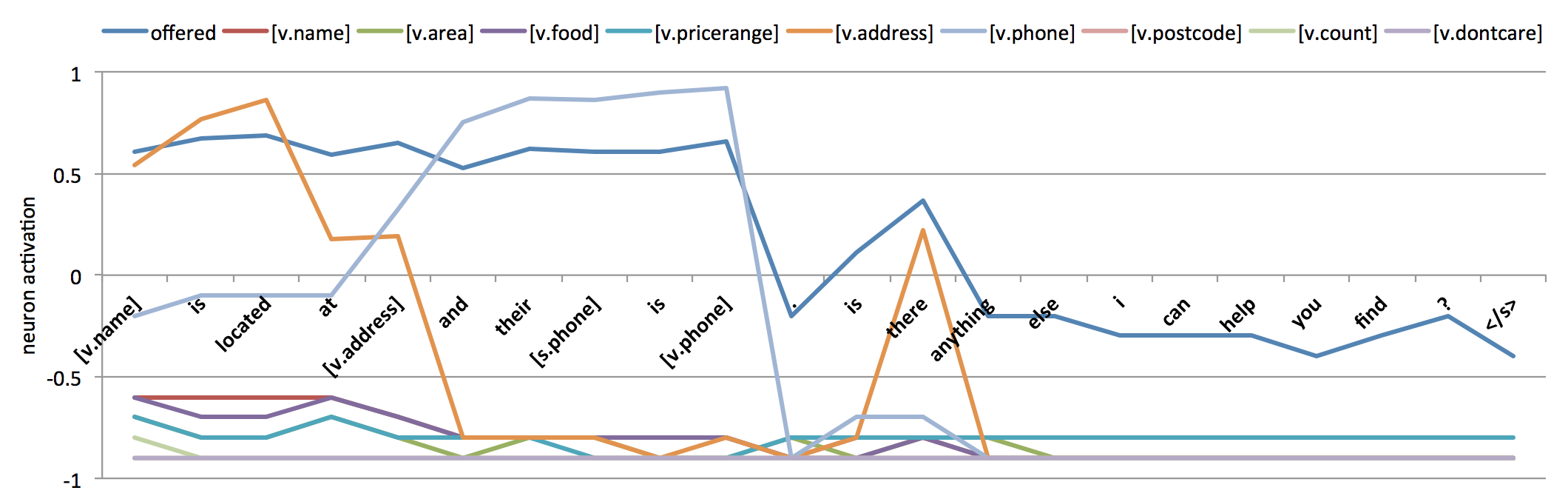}}\ \subfloat[\label{fig:snap2}]{} \\[\topskip]
\vspace{-4mm}
\caption{Two example responses generated from the hybrid model trained with snapshot and attention. Each line represents a neuron that detects a particular snapshot event.}
\label{fig:snapex}
\vspace{-4mm}
\end{figure*}

\noindent {\bf Results} \hspace{2mm} Table~\ref{lab:exp} shows the evaluation results.
The numbers to the left and right of each table cell are the same model trained w/o and w/ snapshot learning.
The first observation is that snapshot learning consistently improves on most metrics regardless of the model architecture.
This is especially true for BLEU scores.
We think this may be attributed to the more discriminative conditioning vector learned through the snapshot method, which makes the learning of the conditional LM easier.

In the first block {\it belief state representation}, we compare the effect of two different belief representations.
As can be seen, using a succinct representation is better ({\it summary}$>${\it full}) because the identity of each categorical value in the belief state does not help when the generation decisions are done in skeletal form.
In fact, the full belief state representation may encourage the model to learn incorrect co-adaptation among features when the data is scarce.

In the {\it conditional architecture} block, we compare the three different conditional generation architectures as described in section~\ref{sssec:condgen}.
This result shows that the language model type ({\it lm}) and memory type ({\it mem}) networks perform better in terms of BLEU score and slot matching rate, while the hybrid type ({\it hybrid}) networks achieve higher task success.
This is probably due to the degree of separation between the LM and conditioning vector: a coupling approach ({\it lm, mem}) sacrifices the conditioning vector but learns a better LM and higher BLEU; while a complete separation ({\it hybrid}) learns a better conditioning vector and offers a higher task success.

Lastly, in the {\it attention-based model} block we train the three architectures with the attention mechanism and compare them again.
Firstly, the characteristics of the three models we observed above also hold for attention-based models.
Secondly, we found that the attention mechanism improves all the three architectures on task success rate but not BLEU scores. 
This is probably due to the limitations of using n-gram based metrics like BLEU to evaluate the generation quality~\cite{Stent05evaluatingevaluation}.

\begin{table}[t]
  \centering
  \Scale[0.95]{\begin{tabular}{lccc}
    \toprule
    Model	&	$\mathrm{\mathbf{i}}_j$	&	$\mathrm{\mathbf{f}}_j$	&	$\mathrm{\mathbf{r}}_j / \mathrm{\mathbf{o}}_j$\\
    \midrule
    %lm, summary  		& 	0.616	&	0.645	& 	0.525   	&	-\\
    %mem, summary		&	0.681	&	0.485	&	0.519 	&	0.339\\
    hybrid, full		&	0.567	&	0.502	&	0.405 	\\
    %\midrule
    hybrid, summary	&	0.539	&	0.540	&	0.428	\\
    + att. 			&	0.540	&	0.559	&	0.459	\\
    \bottomrule
  \end{tabular}}
  \caption{Average activation of gates on test set.}
    \label{tab:gate}
    \vspace{-5mm}
\end{table}

\section{Model Analysis}\label{sec:analysis}

\noindent{\bf Gate Activations} \hspace{2mm} We first studied the average activation of each individual gate in the models by averaging them when running generation on the test set.
We analysed the {\it hybrid} models because their reading gate to output gate activation ratio ($\mathrm{\mathbf{r}}_j / \mathrm{\mathbf{o}}_j$) shows clear tradeoff between the LM and the conditioning vector components.
As can be seen in Table~\ref{tab:gate}, we found that the average forget gate activations ($\mathrm{\mathbf{f}}_j$) and the ratio of the reading gate to the output gate activation ($\mathrm{\mathbf{r}}_j / \mathrm{\mathbf{o}}_j$) have strong correlations to performance:
a better performance ({\it row 3}$>${\it row 2}$>${\it row 1}) seems to come from models that can learn a longer word dependency (higher forget gate $\mathrm{\mathbf{f}}_{t}$ activations) and a better conditioning vector (therefore higher reading to output gate ratio $\mathrm{\mathbf{r}}_j / \mathrm{\mathbf{o}}_j$).

\noindent {\bf Learned Attention} \hspace{2mm} We have visualised the learned attention heat map of models trained with and without snapshot learning in Figure~\ref{fig:att}. 
The attention is on both the informable slot trackers (first three columns) and the requestable slot trackers (the other columns).
We found that the model trained with snapshot learning (Figure~\ref{fig:att2}) seems to produce a more accurate and discriminative attention heat map comparing to the one trained without it (Figure~\ref{fig:att1}).
This may contribute to the better performance achieved by the snapshot learning approach.

\noindent {\bf Snapshot Neurons} \hspace{2mm} As mentioned earlier, snapshot learning forces a subspace of the conditioning vector $\mathrm{\mathbf{\hat{m}}}_{t}^{j}$ to become discriminative and interpretable. 
Two example generated sentences together with the snapshot neuron activations are shown in Figure~\ref{fig:snapex}.
As can be seen, when generating words one by one, the neuron activations were changing to detect different events they were assigned by the snapshot training signals: e.g. in Figure~\ref{fig:snap2} the {\it light blue} and {\it orange} neurons switched their domination role when the token {\it [v.address]} was generated; 
the {\it offered} neuron is in a high activation state in Figure~\ref{fig:snap2} because the system was offering a venue, while in Figure~\ref{fig:snap1} it is not activated because the system was still helping the user to find a venue.
More examples can be found in the Appendix.

\section{Conclusion and Future Work}\label{sec:conclusion}

This paper has investigated different conditional generation architectures and a novel method called snapshot learning to improve response generation in a neural dialogue system framework.
The results showed three major findings. Firstly, although the {\it hybrid type} model did not rank highest on all metrics, it is nevertheless preferred because it achieved the highest task success and also it provided more interpretable results. 
Secondly, snapshot learning provided gains on virtually all metrics regardless of the architecture used.
The analysis suggested that the benefit of snapshot learning mainly comes from the more discriminative and robust subspace representation learned from the heuristically labelled companion signals, which in turn facilitates optimisation of the final target objective.
Lastly, the results suggested that by making a complex system more interpretable at different levels not only helps our understanding but also leads to the highest success rates.

However, there is still much work left to do. This work focused on conditional generation architectures and snapshot learning in the scenario of generating dialogue responses. 
It would be very helpful if the same comparison could be conducted in other application domains such as machine translation or image caption generation so that a wider view of the effectiveness of these approaches can be assessed.

\section*{Acknowledgments}
Tsung-Hsien Wen and David Vandyke are supported by Toshiba Research Europe Ltd, Cambridge Research Laboratory.

\bibliography{emnlp2016}
\bibliographystyle{emnlp2016}

\newpage
\onecolumn

\section*{Appendix: More snapshot neuron visualisation} \label{app:snapshot}

\begin{figure*}[h]
\subfloat{\centerline{\includegraphics[width=160mm]{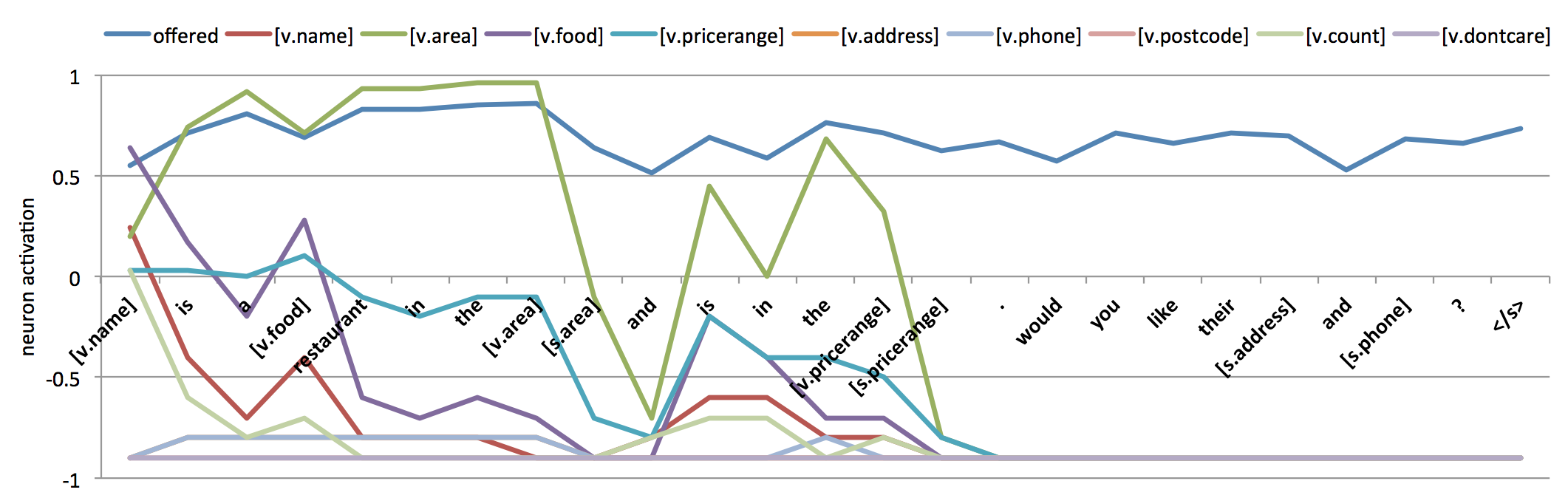}}\label{app:snap3}} \hfill
\subfloat{\centerline{\includegraphics[width=160mm]{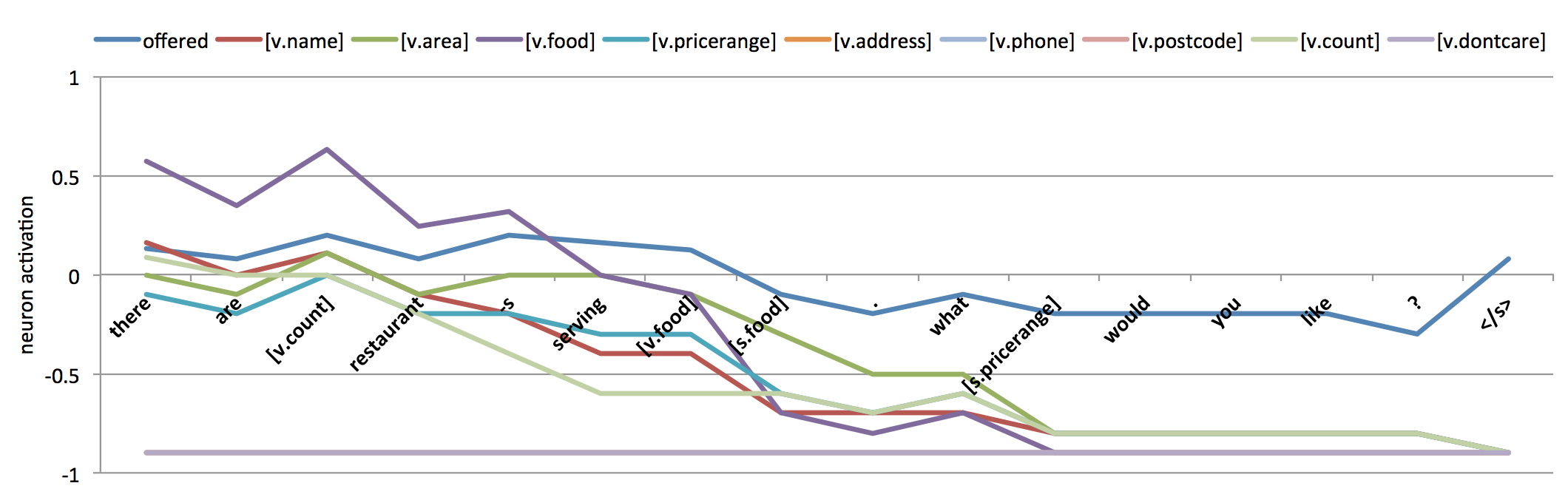}}\label{app:snap4}} \hfill
\subfloat{\centerline{\includegraphics[width=160mm]{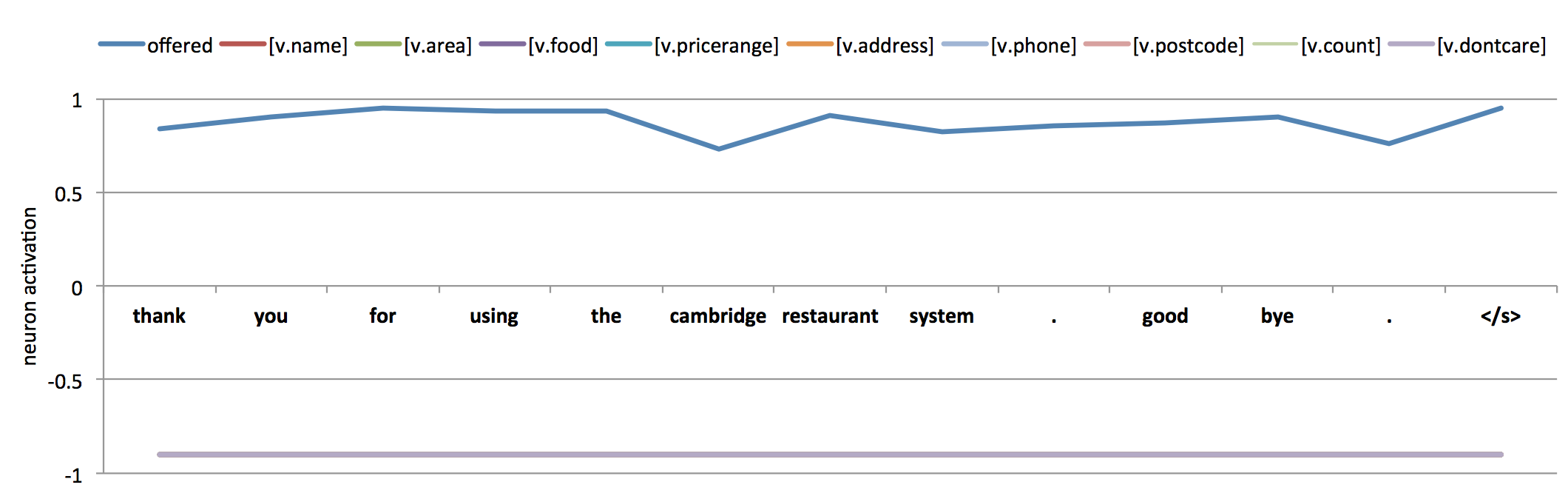}}\label{app:snap5}} 
\caption{More example visualisation of snapshot neurons and different generated responses.}
\centering
\label{app:snapshot}
\end{figure*}

\end{document}